\documentclass[10pt,twocolumn,letterpaper]{article}

\usepackage{graphicx}
\usepackage{booktabs}
\usepackage{multirow}
\usepackage[table]{xcolor}
\usepackage{textcomp}
\definecolor{trainrow}{RGB}{245,245,245}
\definecolor{bluerow}{RGB}{240,245,255}

\usepackage[pagenumbers]{wacv} 

\usepackage[most]{tcolorbox}
\usepackage{caption}
\usepackage{cuted}
\usepackage{dblfloatfix}
\usepackage{tcolorbox}
\tcbuselibrary{skins}
\usepackage[table]{xcolor}

\definecolor{grayrow}{RGB}{245,245,245}
\definecolor{baselinegreen}{RGB}{244,252,244}

\definecolor{orangrow}{RGB}{255,250,225}

\definecolor{greenrow}{RGB}{225,245,225}

\definecolor{baselinegreen}{RGB}{240,250,240}   
\definecolor{oursgreen}{RGB}{215,240,215}        

\definecolor{wacvblue}{rgb}{0.21,0.49,0.74}
\usepackage[pagebackref,breaklinks,colorlinks,allcolors=wacvblue]{hyperref}

\def\wacvPaperID{1733} 
\def\confName{WACV}
\def\confYear{2027}

\title{ATLAS: Learning to Adapt LVLM Attention for Reasoning Video Segmentation}

\author{
Hamidreza Dastmalchi$^{1}$ \quad
Ali Cheraghian$^{2}$ \quad
Aijun An$^{1}$ \quad
Abdelwahed Khamis$^{3}$ \quad
Lars Petersson$^{3}$\\[4pt]
$^{1}$York University \quad
$^{2}$Macquarie University \quad
$^{3}$CSIRO Data61\\[2pt]
{\tt\small \{hrd, aan\}@yorku.ca, ali.cheraghian@mq.edu.au,}\\
{\tt\small \{abdelwahed.khamis, lars.petersson\}@data61.csiro.au}
}

\begin{document}
\maketitle
\begin{abstract}
Reasoning Video Object Segmentation (RVOS) aims to localize and segment objects referred to by natural-language expressions throughout a video. Although large vision-language models (LVLMs) implicitly encode localization cues in their attention maps, these attentions are often noisy, diffuse, and scattered across layers, heads, and modalities, limiting their direct use for segmentation. Existing methods typically rely on handcrafted attention-selection heuristics or computationally expensive attention refinement. We propose ATLAS, an attention adaptation framework that learns to transform raw LVLM attentions into object-centric localization cues. ATLAS treats attentions from all layers, heads, and modalities as structured features, combining high-resolution frame-level attentions that capture spatial information with low-resolution video-level attentions that encode temporal context. A lightweight U-Net equipped with temporal attention and FiLM-based modulation fuses these complementary representations to generate coarse segmentation heatmaps, which are propagated across the video using SAM2. Extensive experiments on challenging referring and reasoning video segmentation benchmarks show that ATLAS outperforms existing frozen-backbone methods. Code and models are available at \url{https://hamidreza-136.github.io/atlas/}.

\end{abstract}
    
\section{Introduction}

Reasoning Video Object Segmentation (RVOS) aims to localize and segment objects referred to by natural-language expressions throughout a video~\cite{reasonvos-videolisa, VISA, VILLA, OnlineRVOS}. 
Unlike conventional video object segmentation, RVOS requires joint reasoning over language, visual content, and temporal context, making Large Vision-Language Models (LVLMs) a natural fit for the task. Recent advances in LVLMs~\cite{LLaVA, LLaVA1.5, internvl, internvl3, qwen2, qwen3, qwen25vl, oryx} have significantly improved multimodal reasoning and reliability~\cite{cipher, cheraghian2026beyond, TTH}, enabling robust understanding of complex expressions and establishing them as a promising backbone for RVOS.

\begin{figure}
    \centering
    \includegraphics[width=\linewidth]{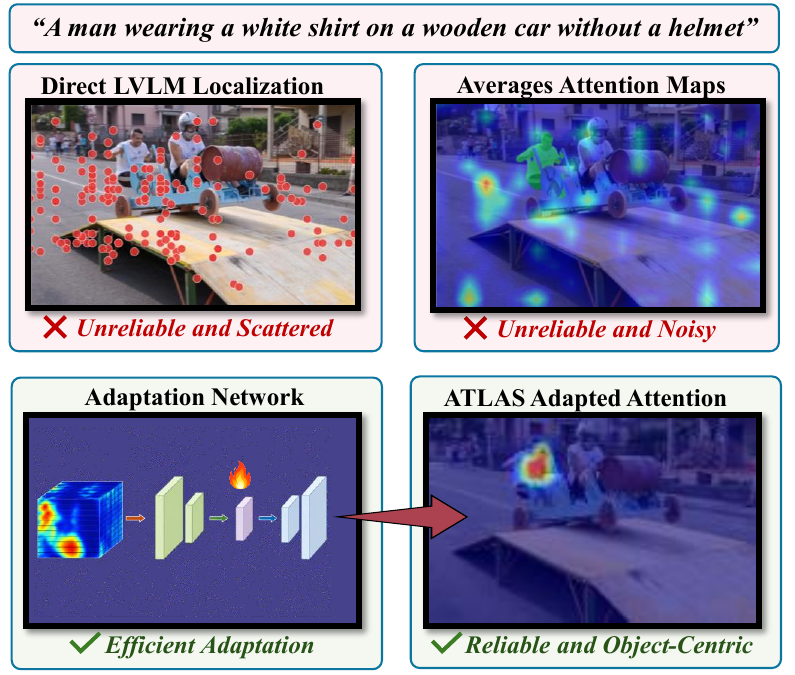}
    \caption{Motivation of ATLAS. Directly querying an LVLM for object coordinates produces inconsistent localization, while naively averaging attention maps across layers and heads yields diffuse and noisy activations. ATLAS instead learns to adapt the complete stack of attention maps into object-centric localization cues using a lightweight adaptation network.}
    \label{fig:motivation}
    \vspace{-0.3cm}
\end{figure}

Recent methods have addressed RVOS by jointly adapting LVLMs with segmentation foundation models, such as SAM \cite{SAM, SAM2}, through end-to-end fine-tuning \cite{reasonvos-videolisa, VILLA, VRS-HQ, GLUS}. While effective, these approaches require jointly optimizing multiple large foundation models, leading to substantial computational cost and dependence on large-scale annotated training data. Moreover, the learned adaptations are often model-specific, limiting their ability to generalize across different LVLMs and segmentation models.

As an efficient alternative, one can directly query a frozen LVLM to predict the $(x,y)$ image coordinates of the referred object. However, LVLMs are trained for general text generation rather than precise coordinate prediction. As illustrated in the top-left panel of Figure~\ref{fig:motivation}, repeatedly querying Qwen2.5-VL~\cite{qwen25vl} 200 times with a sampling temperature of 0.7 produces highly inconsistent coordinate estimates that are widely scattered across the image, making direct localization unreliable.

Nevertheless, recent studies have shown that the visual-token attentions of frozen LVLMs implicitly encode object localization information, enabling visual grounding~\cite{LOC-HEAD,DecAF,vl-sam,TAM, steerseg}. However, effectively extracting these localization cues remains challenging. As illustrated in the top-right panel of Figure~\ref{fig:motivation}, averaging attentions across layers and heads produces diffuse, noisy activation maps poorly aligned with object boundaries. Existing training-free approaches either manually select a small subset of attention heads believed to encode localization cues \cite{LOC-HEAD} or refine attention maps through repeated LVLM inference with contrastive prompting \cite{DecAF}, leaving much of the localization information distributed across layers and heads underutilized. Recent work has shown that a learned decoder over the full attention stack can recover this information for image grounding \cite{F-LMM}. Reasoning video segmentation, however, introduces a fundamentally different challenge: the model must jointly infer where the referred object is and when it becomes relevant over time. We therefore investigate how the full multimodal attention stack can be adapted to capture both spatial grounding and temporal relevance.

To address these limitations, we propose \textbf{ATLAS} (\textbf{A}ttention \textbf{T}ransformation for \textbf{L}ocalization through \textbf{A}daptive \textbf{S}egmentation), which leverages the complete visual attention representation from the middle-to-last layers across all attention heads as a stack of spatial feature maps, rather than treating attention maps as localization outputs to be manually selected or refined. As illustrated at the bottom of Figure~\ref{fig:motivation}, a lightweight attention adaptation network transforms these rich attention features into object-centric localization cues. The resulting adapted attention maps are then used to guide a frozen SAM2~\cite{SAM2} model, producing accurate and temporally consistent video object segmentation.

Given a video-expression pair, ATLAS first samples a set of representative keyframes and extracts visual attention maps from a frozen LVLM. Specifically, it obtains high-resolution frame-level attentions that capture \textbf{where} the referred object is located and low-resolution video-level attentions that encode \textbf{when} the object becomes relevant. The stacked attention features are processed by a lightweight encoder-decoder U-Net, while a temporal attention module aggregates complementary information across the sampled keyframes. The resulting temporal representation modulates the U-Net bottleneck through \textbf{Feature-wise Linear Modulation (FiLM)}, enabling the decoder to produce temporally-aware, object-centric localization maps. These adapted attention maps are subsequently used to prompt a frozen SAM2 model, yielding accurate and temporally consistent segmentation masks throughout the video. Overall, our contributions are summarized as follows:

\begin{itemize}
    \item We propose \textbf{ATLAS}, an attention adaptation framework for reasoning video segmentation that transforms complete LVLM attention representations into object-centric localization cues, eliminating the need for handcrafted attention selection or costly refinement.

    \item We introduce a lightweight attention adaptation network that jointly exploits frame-level attentions and video-level attentions, integrating temporal cues into spatial localization to model both \emph{where} the target object is located and \emph{when} it becomes relevant.
    
    \item We achieve state-of-the-art performance on reasoning video segmentation among methods with frozen backbones, using a lightweight supervised adapter while keeping the LVLM and segmentation model fully frozen.

\end{itemize}
\section{Related Work}

\noindent\textbf{Reasoning Video Object Segmentation.} Reasoning video object segmentation (RVOS)  extends referring video object segmentation~\cite{ReferFormer,SOC, MTTR, referdino} by requiring semantic reasoning over complex language expressions and visual-temporal context~\cite{reasonvos-videolisa, VISA, VILLA, VRS-HQ, pixellm}.  Early approaches primarily relied on multimodal fusion architectures that jointly encoded visual and linguistic representations to predict segmentation masks ~\cite{Refer-Youtube-VOS, ReferFormer, SOC, MTTR}. More recent methods leverage large-scale pretrained LVLMs together with segmentation foundation models to improve semantic grounding and generalization~\cite{LISA,VISA,reasonvos-videolisa,VILLA}. Despite this progress, RVOS remains challenging because accurate segmentation requires not only semantic understanding but also precise spatial localization and reasoning over temporal dynamics.

\vspace{0.1cm}
\noindent\textbf{Large Vision-Language Models for Visual Grounding.}
Large Vision-Language Models (LVLMs), such as LLaVA~\cite{LLaVA, LLaVA1.5}, Qwen2-VL~\cite{qwen2}, and Qwen2.5-VL~\cite{qwen25vl}, have demonstrated strong capabilities in multimodal reasoning and visual grounding. Recent studies have shown that LVLMs implicitly encode object localization cues in their internal representations, particularly in the attention maps of visual tokens. This observation has motivated a growing body of work that repurposes frozen LVLMs for visual grounding and segmentation by extracting attention maps or other internal representations as localization cues~\cite{DecAF,LOC-HEAD}. However, these representations are often coarse and poorly aligned with object boundaries, motivating better approaches for exploiting LVLM attention.
\begin{figure*}[!t]
    \centering
    \includegraphics[width=\linewidth]{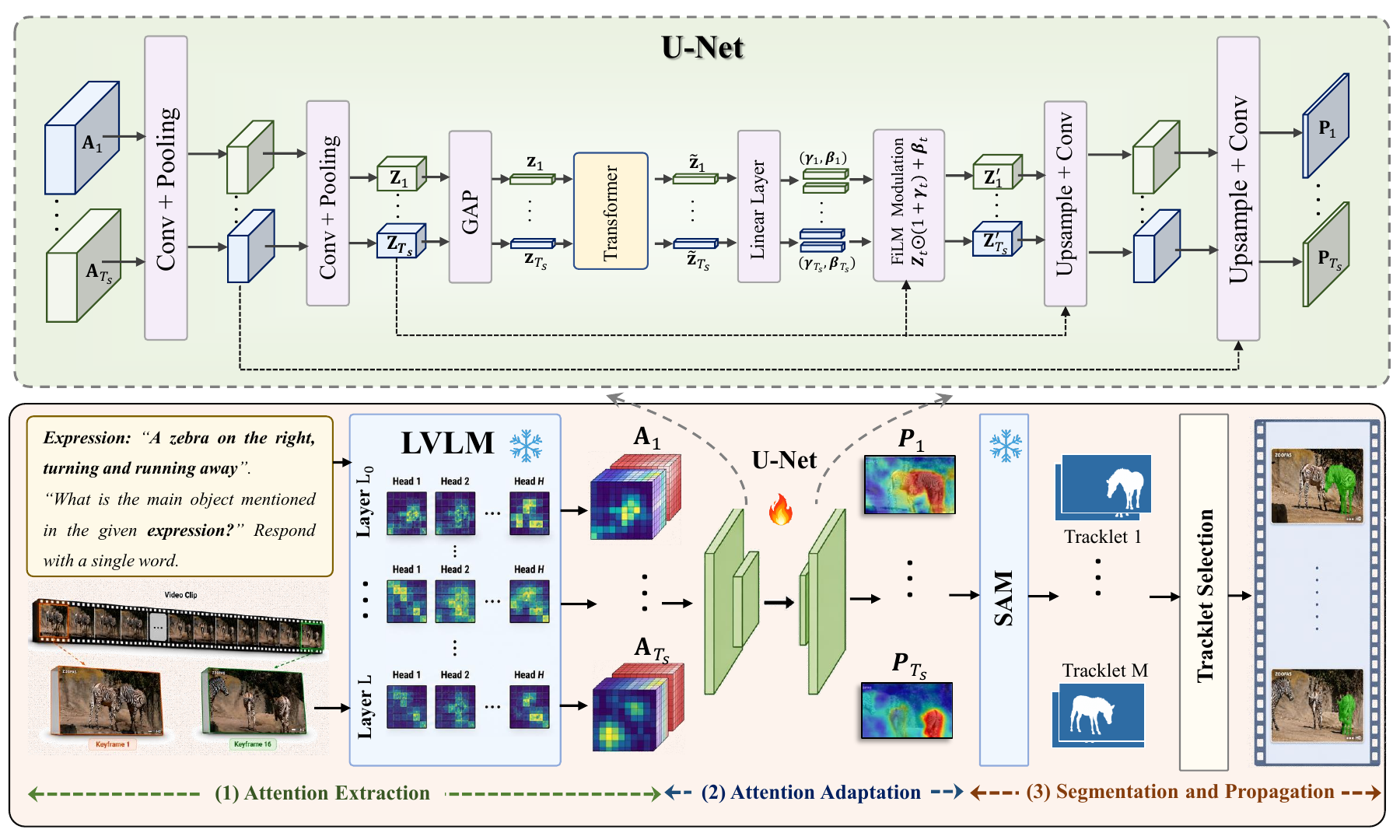}
    \caption{\textbf{Overview of ATLAS.} A frozen LVLM extracts visual attentions from sampled keyframes, which are transformed into object-centric localization maps by a lightweight attention adaptation network. The adapted localization maps guide a frozen SAM2 model to generate and propagate candidate segmentation tracklets, from which the most reliable tracklet is selected as the final segmentation result.}
    \label{fig:pipeline}
\end{figure*}

\vspace{0.1cm}
\noindent\textbf{Attention-Based Localization from LVLMs.}
Recent studies exploit LVLM attention maps as localization cues for visual
grounding and segmentation \cite{TAM, vl-sam, LOC-HEAD, DecAF, F-LMM}.
TAM~\cite{TAM} utilizes attention rollout~\cite{rollout} to obtain object
localization cues from LVLM attentions, whereas VL-SAM~\cite{vl-sam}
enhances these cues using SAM~\cite{SAM}. Both methods require enumerating
object categories during LVLM decoding to detect all objects in an image.
Loc-Head~\cite{LOC-HEAD} identifies a small set of \emph{localization
heads} that consistently focus on referred objects and aggregates their
attentions to estimate object locations. Others, such as DecAF~\cite{DecAF},
improve localization by refining attention maps through contrastive
foreground-background prompting and multi-stage attention processing.
These training-free approaches either discard informative attentions or
require repeated inference. F-LMM~\cite{F-LMM} instead learns to decode the
attention representation, training a small CNN over the stacked
layer-head attention of a frozen LMM to predict grounding masks for
images, which a trainable SAM-based refiner then optimizes. ATLAS shares this per-frame primitive but targets reasoning video segmentation, where
localization must model \emph{when} the referred object becomes relevant
in addition to \emph{where} it is located.

\section{Method}

\subsection{Method Overview}
Given a video
\(
\mathcal{V}=\{\mathbf{I}_t\}_{t=1}^{T}
\)
and a reasoning expression \(\mathcal{E}\), the goal is to predict a sequence of binary masks
\(
\mathcal{M}=\{\mathbf{M}_t\}_{t=1}^{T},
\)
where \(\mathbf{I}_t \in \mathbb{R}^{R \times W \times 3}\) is the RGB frame at time step \(t\), with spatial resolution \(R \times W\) (height \(\times\) width), and \(\mathbf{M}_t \in \{0,1\}^{R \times W}\) denotes the segmentation mask of the referred object.
Our framework consists of three stages (depicted in Figure~\ref{fig:pipeline}).

\noindent\textbf{(1) Attention Extraction:}
Given a video-expression pair, a frozen LVLM extracts frame-level and video-level visual attention maps from a set of sparsely sampled keyframes.

\noindent\textbf{(2) Attention Adaptation:}
ATLAS transforms the complete stack of attention maps from the middle-to-last layers across all attention heads into object-centric localization maps using a lightweight attention adaptation network.

\noindent\textbf{(3) Segmentation and Propagation:}
The adapted attention maps are converted into point prompts that guide a frozen SAM2 model to segment the sampled keyframes and propagate the masks to the remaining frames.

\subsection{Attention Extraction}
\vspace{0.1cm}

Given a video-expression pair {$(\mathcal{V}, \mathcal{E})$}, we first uniformly sample a fixed set of \(T_s\) frames from the input video, denoted by
\(
\mathcal{V}_s=\{\mathbf{I}_{t_i}\}_{i=1}^{T_s},
\)
where \(t_i\) is the index of the \(i\)-th sampled frame. We augment the expression with a fixed object extraction prompt (provided in Appendix~\ref{app:prompt}) and query a frozen LVLM to generate a single response token. We then extract the attention of this token to all visual tokens across transformer layers and attention heads.

For frame-level attention, the LVLM is queried independently for each sampled frame, resulting in \(T_s\) separate forward passes with a high visual-token budget (e.g., 1024 visual tokens). In contrast, video-level attention is obtained from a single forward pass in which the entire sampled frame sequence \(\mathcal{V}_s\) is processed jointly using a smaller visual-token budget (e.g., 64 visual tokens), providing complementary temporal cues across frames.

For each keyframe $\mathbf{I}_{t_k}$, let
\(\mathbf{A}^{(f)}_{k,\ell,h}\in\mathbb{R}^{R_f\times W_f}\)
denote the layer-$\ell$, head-$h$ attention from the generated response token to the visual tokens, reshaped to the LVLM’s spatial token grid.
We collect and channel-stack attention maps from all heads across layers \(\ell=L_0,\ldots,L\), where \(L_0\) is the first selected layer. Following prior observations that middle-to-late layers provide stronger semantic localization cues than early layers \cite{DecAF}, we set \(L_0\) to the middle transformer layer.

\begin{equation}
\mathbf{A}^{(f)}_{k}
=
\operatorname{Stack}_{\ell,h}
\!\left(
\mathbf{A}^{(f)}_{k,\ell,h}
\right)
\in
\mathbb{R}^{C \times R_f \times W_f},
\end{equation}

\noindent where \(C=(L-L_0+1)\times H\) denotes the total number of stacked attention maps.

Similarly, let
\(
\mathbf{A}^{(v)}_{k,\ell,h}
\in
\mathbb{R}^{R_v \times W_v}
\)
denote the corresponding attention map extracted from the video branch, where \(R_v\) and \(W_v\) denote the height and width of the video-attention maps, respectively. Stacking attentions from the same set of layers and heads yields:

\begin{equation}
\mathbf{A}^{(v)}_{k}
=
\operatorname{Stack}_{\ell,h}
\!\left(
\mathbf{A}^{(v)}_{k,\ell,h}
\right)
\in
\mathbb{R}^{C \times R_v \times W_v},
\end{equation}

The video attentions are then aligned to the frame-attention resolution using bilinear interpolation and concatenated channel-wise:

\begin{equation}
\mathbf{A}_k=
\left[
\mathbf{A}^{(f)}_k;
\operatorname{Resize}(\mathbf{A}^{(v)}_k)
\right]
\in
\mathbb{R}^{2C \times R_f \times W_f}.
\end{equation}

The resulting attention tensor $\mathbf{A}_k$, corresponding to the $k$-th sampled keyframe, serves as the input to the proposed attention adaptation network.

\vspace{0.1cm}
\subsection{Attention Adaptation}

The second stage transforms the extracted attention representations into object-centric localization maps using a two-stage lightweight U-Net. Each stage consists of two convolutional layers, and the corresponding encoder and decoder stages are connected through standard skip connections.

Given a stacked attention tensor
\(
\mathbf{A}_k\in\mathbb{R}^{2C\times R_f\times W_f}
\) for \(k\)-th keyframe,
the encoder progressively aggregates spatial information through convolution and pooling operations to produce a latent feature tensor

\begin{equation}
\mathbf{Z}_k
=
\mathrm{Encoder}(\mathbf{A}_k)
\in
\mathbb{R}^{d\times r\times w},
\;\; k=1,\ldots,T_s.
\end{equation}

\noindent where \(r\) and \(w\) denote the spatial dimensions of the bottleneck feature map, and \(d\) denotes its channel dimension.

The bottleneck feature maps are then  globally pooled into latent vectors using global average pooling (GAP):

\begin{equation}
\mathbf{z}_k
=
\mathrm{GAP}(\mathbf{Z}_k)
\in
\mathbb{R}^{d},
\qquad k=1,\ldots,T_s.
\end{equation}

These latent vectors are subsequently processed by a lightweight two-layer Transformer encoder to model temporal dependencies across the sampled keyframes:

\begin{equation}
\tilde{\mathbf{z}}_1,\ldots,\tilde{\mathbf{z}}_{T_s}
=
\mathrm{Transformer}
(\mathbf{z}_1,\ldots,\mathbf{z}_{T_s}).
\end{equation}

 Temporal context is injected into the spatial bottleneck features through \textbf{Feature-wise Linear Modulation (FiLM)} \cite{FiLM}, which provides a lightweight mechanism for conditioning the decoder on temporally contextualized representations while preserving the spatial feature structure. Specifically, the contextualized tokens are projected into FiLM parameters through two learnable linear transformations:
 \vspace{-0.2cm}
\begin{equation}
\begin{aligned}
\boldsymbol{\gamma}_k
&=
\mathbf{W}_{\gamma}\tilde{\mathbf{z}}_k
+
\mathbf{b}_{\gamma}, \qquad
k=1,\ldots,T_s.\\
\boldsymbol{\beta}_k
&=
\mathbf{W}_{\beta}\tilde{\mathbf{z}}_k
+
\mathbf{b}_{\beta},
\qquad
k=1,\ldots,T_s.
\end{aligned}
\end{equation}

\noindent where \(\mathbf{W}_{\gamma}, \mathbf{W}_{\beta}\in\mathbb{R}^{d\times d}\) and
\(\mathbf{b}_{\gamma}, \mathbf{b}_{\beta}\in\mathbb{R}^{d}\) are learnable projection parameters, and
\(\boldsymbol{\gamma}_k,\boldsymbol{\beta}_k\in\mathbb{R}^{d}\) denote the feature-wise scaling and shifting coefficients, respectively. These coefficients are broadcast across the spatial dimensions and used to modulate the bottleneck features via FiLM:

\begin{equation}
\mathbf{Z}'_k
=
\mathbf{Z}_k
\odot
(1+\boldsymbol{\gamma}_k)
+
\boldsymbol{\beta}_k, \qquad
k=1,\ldots,T_s.
\end{equation}

By conditioning the bottleneck features on temporally contextualized representations, the decoder exploits temporal context to produce more consistent object localization across keyframes. The modulated features are decoded into a localization probability map,

\begin{equation}
\mathbf{P}_k
=
\mathrm{Decoder}
(\mathbf{Z}'_k)
\in
[0,1]^{R_f\times W_f}, \;\;
k=1,\ldots,T_s.
\end{equation}

\noindent where \(\mathbf{P}_k\) denotes the predicted localization probability map for keyframe \(k\).

The attention adaptation network is trained using a combination of Binary Cross-Entropy (BCE) and Soft Dice losses \cite{Dice}:

\begin{equation}
\mathcal{L}
=
\mathcal{L}_{\mathrm{BCE}}
+
\mathcal{L}_{\mathrm{Dice}},
\end{equation}

\noindent where
\(\mathcal{L}_{\mathrm{BCE}}\)
is the pixel-wise binary cross-entropy between the predicted probability map
\(\mathbf{P}_k\)
and the ground-truth mask
\(\mathbf{M}_{t_k}\),
and
\(\mathcal{L}_{\mathrm{Dice}}\)
is the corresponding Soft Dice loss.

Notably, only the attention adaptation network is optimized, while the LVLM and SAM2 remain frozen. It learns to transform the complete attention representation into object-centric localization maps, eliminating the need for handcrafted head selection or attention refinement.

\subsection{Segmentation and Propagation}
\label{sec: segmentation}
The third stage converts the predicted localization maps into temporally
consistent object masks using a frozen SAM2 model. For each keyframe, a
positive point prompt is selected as the maximum-response location of the
predicted localization map,
\begin{equation}
\mathbf{p}_k
=
\arg\max_{(u,v)}
\mathbf{P}_k(u,v)  \qquad
k=1,\ldots,T_s.
\end{equation}
The resulting point prompts are provided to SAM2 to generate object masks
on the sampled keyframes. Several keyframes may be seeded on the same
object, so redundant seeds are removed before propagation. Spatial
non-maximum suppression across keyframes, as used in DecAF~\cite{DecAF},
is insufficient here: it compares masks at different time steps, so if
the object moves, its masks at two keyframes do not overlap and both
seeds survive even though they refer to the same object. Instead, we test
whether two seeds correspond to the same object by propagating one to the
other's keyframe. Starting from the seed at the earliest keyframe, we
propagate its mask through the video; at every later keyframe $k$, we
compare the propagated mask with the seed mask at $k$ and discard the
latter if their IoU exceeds a threshold $\tau$ (we use
$\tau=0.8$), since the object it identifies is
already tracked. We then repeat the procedure from the earliest remaining
seed until none are left. This yields $M$ tracklets corresponding to
distinct object hypotheses, and each duplicate removed also saves a full
propagation pass.

Each tracklet $\boldsymbol{\mathcal{T}}_m$ is downsampled to the attention
resolution and sampled at the $T_s$ keyframes, yielding a tracklet volume
$\mathbf{V}_m \in [0,1]^{T_s \times R_f \times W_f}$ for
$m=1,\ldots,M$. The final tracklet is selected as the one whose spatial
activation pattern best matches the predicted localization maps, averaged
over the sampled keyframes:
\begin{equation}
m^{*}
=
\arg\max_{m}\;
\frac{1}{T_s}\sum_{k=1}^{T_s}
\rho\!\left(\mathbf{V}_{m,k},\, \mathbf{P}_k\right),
\label{eq:eq12}
\end{equation}
where $\mathbf{V}_{m,k}\in[0,1]^{R_f\times W_f}$ denotes the $k$-th
keyframe slice of $\mathbf{V}_m$, and $\rho(\cdot,\cdot)$ is the Pearson
correlation over the spatial locations of a single keyframe. Pearson
correlation measures similarity between spatial activation patterns while
being invariant to linear scaling and shifts in magnitude, making it robust
to confidence differences between predicted localization maps and binary
tracklet masks. Computing it per keyframe and averaging normalizes each
keyframe's contribution, so frames where the object is small or briefly
visible are weighted equally to those where it is large and confidently
localized. If the maximum correlation is below $0.1$, we return an empty
mask, indicating that no reliable tracklet was found. Otherwise, the final
segmentation is given by $\boldsymbol{\mathcal{T}}_{m^{*}}$, providing
temporally consistent object masks for the full video.

\section{Experiments}
\label{sec:experiment}

\textbf{Datasets.} We evaluate ATLAS on five referring video object segmentation benchmarks, including three conventional datasets: Ref-DAVIS~\cite{Davis}, Ref-YouTube-VOS~\cite{Refer-Youtube-VOS}, and MeViS~\cite{Mevis}, together with two reasoning-oriented benchmarks: ReasonVOS~\cite{reasonvos-videolisa} and ReVOS~\cite{VISA}. While the conventional benchmarks primarily evaluate referring object segmentation, ReasonVOS and ReVOS contain more challenging expressions that require multi-step reasoning, compositional understanding, and temporal reasoning.

\vspace{0.15cm}
\noindent \textbf{Evaluation Metrics.} Following  \cite{DecAF}, we report the standard video object segmentation metrics: region similarity (\(\mathcal{J}\)), contour accuracy (\(\mathcal{F}\)), and their average (\(\mathcal{J}\&\mathcal{F}\)). Specifically, \(\mathcal{J}\) measures the intersection-over-union (IoU) between the predicted and ground-truth masks, while \(\mathcal{F}\) evaluates boundary accuracy by comparing the predicted and ground-truth object contours. The combined \(\mathcal{J}\&\mathcal{F}\) score provides an overall measure of segmentation quality by jointly assessing region overlap and boundary precision.

\vspace{0.15cm}
\noindent\textbf{Models.} We use publicly available pretrained LVLM checkpoints, including LLaVA-OV-7B~\cite{llava-ov}, InternVL3-8B~\cite{internvl3}, Qwen2-VL-7B~\cite{qwen2}, and Qwen2.5-VL-7B~\cite{qwen25vl}. For mask propagation, we adopt the pretrained SAM2-Hiera-Large~\cite{SAM2} segmentation model.

\vspace{0.15cm}
\noindent\textbf{Implementation Details.}
The attention adaptation network is a lightweight U-Net with two encoder and two decoder stages, where each stage consists of two convolutional layers. The bottleneck feature has dimension $d=512$ and is conditioned via FiLM \cite{FiLM} using a two-layer temporal Transformer with four attention heads. For attention extraction, we use a maximum visual token budget of {1024} for the frame branch to obtain high-resolution spatial attentions and {64} for the video branch to capture complementary temporal cues. We train the network exclusively on the Ref-Youtube-VOS~\cite{Refer-Youtube-VOS} training split, demonstrating strong generalization to unseen reasoning video segmentation benchmarks. The network is optimized using AdamW with a batch size of 8, an initial learning rate of $1\times10^{-4}$ decayed linearly to $1\times10^{-5}$, and is trained for 20 epochs. Training requires approximately 6 GPU hours on a single NVIDIA RTX Pro 6000 GPU. Additional architectural details and training hyperparameters are provided in Appendix~\ref{sec:appendix-imlementation-details}.

\subsection{Results}

\begin{table*}[t]
\centering
\caption{
Comparison of LVLM-based VOS methods on Ref-DAVIS~\cite{Davis}, ReasonVOS~\cite{reasonvos-videolisa}, and ReVOS~\cite{VISA}.  $^\dagger$ Loc-Head and DecAF are training-free; ATLAS trains a lightweight adapter with mask supervision. All three keep the LVLM and segmentation model frozen.
}
\label{tab:comparison2}

\definecolor{trainrow}{RGB}{250,240,240}   
\definecolor{bluerow}{RGB}{240,245,255}    

\definecolor{baselineorange}{RGB}{242,247,255}
\definecolor{oursorange}{RGB}{215,230,250}

\definecolor{baselinepurple}{RGB}{255,252,235}
\definecolor{ourspurple}{RGB}{250,242,195}
\definecolor{baselineteal}{RGB}{252,244,240}
\definecolor{oursteal}{RGB}{245,225,218}

\definecolor{baselinegreen}{RGB}{244,252,244}
\definecolor{oursgreen}{RGB}{225,245,225}

\resizebox{\textwidth}{!}{%
\begin{tabular}{llccccccccccccccc}
\toprule
\multirow{2}{*}{\textbf{Method}} &
\multirow{2}{*}{\textbf{LVLM}}
& \multicolumn{3}{c}{\textbf{Ref-DAVIS}}
& \multicolumn{3}{c}{\textbf{ReasonVOS}}
& \multicolumn{3}{c}{\textbf{ReVOS\textsubscript{(Overall)}}}
& \multicolumn{3}{c}{\textbf{ReVOS\textsubscript{(Referring)}}}
& \multicolumn{3}{c}{\textbf{ReVOS\textsubscript{(Reasoning)}}} \\

\cmidrule(lr){3-5}
\cmidrule(lr){6-8}
\cmidrule(lr){9-11}
\cmidrule(lr){12-14}
\cmidrule(lr){15-17}

&
& $\boldsymbol{\mathcal{J\&F}}$ & $\boldsymbol{\mathcal{J}}$ & $\boldsymbol{\mathcal{F}}$
& $\boldsymbol{\mathcal{J\&F}}$ & $\boldsymbol{\mathcal{J}}$ & $\boldsymbol{\mathcal{F}}$
& $\boldsymbol{\mathcal{J\&F}}$ & $\boldsymbol{\mathcal{J}}$ & $\boldsymbol{\mathcal{F}}$
& $\boldsymbol{\mathcal{J\&F}}$ & $\boldsymbol{\mathcal{J}}$ & $\boldsymbol{\mathcal{F}}$
& $\boldsymbol{\mathcal{J\&F}}$ & $\boldsymbol{\mathcal{J}}$ & $\boldsymbol{\mathcal{F}}$ \\

\midrule
\midrule
\rowcolor{baselineorange}
\multicolumn{17}{c}{\textbf{Approaches with tuned foundation models}} \\
\midrule

\rowcolor{baselineorange}
LISA \cite{LISA}
& LLaVA-7B
& 64.8 & 62.2 & 67.3
& 31.1 & 29.1 & 33.1
& 40.9 & 39.1 & 42.7
& 45.7 & 44.3 & 47.1
& 36.1 & 33.8 & 38.4 \\

\rowcolor{baselineorange}
VISA \cite{VISA}
& ChatUniVi-7B
& 69.4 & 66.3 & 72.5
& -- & -- & --
& 46.9 & 44.9 & 49.0
& 50.9 & 49.2 & 52.6
& 43.0 & 40.6 & 45.4 \\

\rowcolor{baselineorange}
GLUS \cite{GLUS}
& LLaVA-7B
& -- & -- & --
& 49.9 & 47.5 & 52.4
& 54.9 & 52.4 & 57.3
& 58.3 & 56.0 & 60.7
& 51.4 & 48.8 & 53.9 \\

\rowcolor{baselineorange}
VideoLISA \cite{reasonvos-videolisa}
& LLaVA-Phi-3-V
& 68.8 & 64.9 & 72.7
& 47.5 & 45.1 & 49.9
& -- & -- & --
& -- & -- & --
& -- & -- & -- \\

\rowcolor{baselineorange}
VRS-HQ \cite{VRS-HQ}
& ChatUniVi-7B
& 76.0 & 72.6 & 79.4
& -- & -- & --
& 59.1 & 56.6 & 61.6
& 62.1 & 59.8 & 64.5
& 56.1 & 53.5 & 58.7 \\

\rowcolor{baselineorange}
Veason-R1 \cite{Veason-R1}
& Qwen2.5VL-7B
& -- & -- & --
& 59.9 & 56.0 & 63.8
& 61.3 & 58.2 & 64.4
& 63.6 & 60.7 & 66.5
& 59.0 & 55.8 & 62.2 \\

\midrule
\midrule

\rowcolor{baselinegreen}
\multicolumn{17}{c}{\textbf{Approaches with frozen foundation models}} \\
\midrule


\rowcolor{baselinegreen}
Loc-Head \cite{LOC-HEAD}  $^\dagger$
& LLaVA-7B
& 56.3 & 52.1 & 60.5
& 33.6 & 29.3 & 38.0
& 32.5 & 28.2 & 36.9
& 36.9 & 32.5 & 41.3
& 28.1 & 23.8 & 32.5 \\

\rowcolor{baselinegreen}
DecAF \cite{DecAF}  $^\dagger$
& LLaVA-OV-7B
& 59.4 & 54.8 & 64.0
& {52.8} & {49.3} & \textbf{56.3}
& 40.0 & 35.8 & 44.1
& 43.4 & 39.1 & 47.6
& 36.6 & 32.6 & 40.7 \\

\rowcolor{oursgreen}
\textbf{ATLAS}
& LLaVA-OV-7B
& \textbf{68.6} & \textbf{65.9} & \textbf{71.3}
& \textbf{53.0} & \textbf{49.8} & {56.2}
& \textbf{48.0} & \textbf{44.8} & \textbf{51.3}
& \textbf{53.4} & \textbf{50.4} & \textbf{56.4}
& \textbf{43.0} & \textbf{39.5} & \textbf{46.6} \\

\midrule


\rowcolor{baselinegreen}
Loc-Head \cite{LOC-HEAD}  $^\dagger$
& InternVL3-8B
& 66.3 & 62.4 & 70.2
& 44.3 & 41.0 & 47.5 
& 43.7 & 39.9 & 47.5
& 46.7 & 42.9 & 50.6
& 40.7 & 36.9 & 44.5 \\

\rowcolor{baselinegreen}
DecAF \cite{DecAF}  $^\dagger$
& InternVL3-8B
& 62.8 & 56.9 & 68.6
& \textbf{58.9} & \textbf{55.1} & \textbf{62.7}
& 47.4 & 43.7 & 51.2
& 51.7 & 47.9 & 55.5
& 43.2 & 39.5 & 46.8 \\

\rowcolor{oursgreen}
\textbf{ATLAS}
& InternVL3-8B
& \textbf{74.4} & \textbf{71.5} & \textbf{77.4}
& {57.2} & {54.2} & {60.2}
& \textbf{53.5} & \textbf{50.8} & \textbf{56.2}
& \textbf{60.0} & \textbf{57.4} & \textbf{62.6}
& \textbf{47.4} & \textbf{44.5} & \textbf{50.3} \\

\midrule


\rowcolor{baselinegreen}
Loc-Head \cite{LOC-HEAD}  $^\dagger$
& Qwen2VL-7B
& 61.9 & 58.0 & 65.8
& 34.0 & 31.8 & 36.2
& 44.0 & 40.8 & 47.2
& 52.7 & 49.1 & 56.2
& 35.4 & 32.6 & 38.2 \\

\rowcolor{baselinegreen}
DecAF \cite{DecAF}  $^\dagger$
& Qwen2VL-7B
& 64.1 & 59.4 & 68.9
& 52.5 & 49.0 & 56.0
& 45.3 & 41.6 & 49.0
& 52.7 & 48.9 & 56.4
& 37.9 & 34.3 & 41.5 \\

\rowcolor{oursgreen}
\textbf{ATLAS}
& Qwen2VL-7B
& \textbf{75.5} & \textbf{72.1} & \textbf{78.9}
& \textbf{64.4} & \textbf{61.8} & \textbf{67.0}
& \textbf{53.8} & \textbf{50.8} & \textbf{56.9}
& \textbf{59.8} & \textbf{56.9} & \textbf{62.6}
& \textbf{48.3} & \textbf{44.9} & \textbf{51.7} \\

\midrule


\rowcolor{baselinegreen}
Loc-Head \cite{LOC-HEAD}  $^\dagger$
& Qwen2.5VL-7B
& 64.6 & 60.2 & 68.9
& 41.1 & 37.9 & 44.3
& 47.0 & 43.3 & 50.7
& 53.1 & 49.3 & 56.9
& 40.8 & 37.2 & 44.4 \\

 \rowcolor{baselinegreen}
DecAF \cite{DecAF}  $^\dagger$
& Qwen2.5VL-7B
& 75.2 & 70.9 & 79.5
& 63.9 & 60.5 & {67.2}
& 54.2 & 50.1 & 58.2
& 58.7 & 54.8 & 62.6
& 49.7 & 45.4 & 53.9 \\

\rowcolor{oursgreen}
\textbf{ATLAS}
& Qwen2.5VL-7B
& \textbf{80.1} & \textbf{76.8} & \textbf{83.4}
& \textbf{67.4} & \textbf{64.7} & \textbf{70.2}
& \textbf{56.2} & \textbf{53.5} & \textbf{59.0}
& \textbf{61.5} & \textbf{58.9} & \textbf{64.1}
& \textbf{51.0} & \textbf{48.1} & \textbf{53.9} \\
\bottomrule
\end{tabular}}
\end{table*}

\paragraph{Comparison with Existing Methods.}

Table~\ref{tab:comparison2} compares ATLAS with both fully fine-tuned and frozen-backbone LVLM-based video object segmentation methods across five benchmark settings. ATLAS achieves state-of-the-art performance among frozen-backbone approaches.

Compared with previous frozen-backbone methods, namely Loc-Head~\cite{LOC-HEAD} and DecAF~\cite{DecAF}, ATLAS consistently delivers large improvements across different LVLM backbones. The most significant gains are observed with Qwen2-VL-7B, where ATLAS improves the $\mathcal{J}\&\mathcal{F}$ score on ReasonVOS from 52.5 to 64.4 (+11.9) and on ReVOS from 45.3 to 53.8 (+8.5) over DecAF. Similarly, with InternVL3-8B, ATLAS boosts Ref-DAVIS from 62.8 to 74.4 (+11.6) and ReVOS-Referring from 51.7 to 60.0 (+8.3). These consistent gains across diverse LVLMs demonstrate that the proposed attention adaptation module effectively transforms noisy LVLM attentions into accurate object localization cues, leading to substantially improved segmentation performance.

Notably, ATLAS achieves performance comparable to end-to-end fine-tuned approaches despite lightweight training. With Qwen2.5-VL-7B, it achieves the highest $\mathcal{J}\&\mathcal{F}$ on Ref-DAVIS and ReasonVOS, reaching 80.1 and 67.4. Across frozen-backbone settings, ATLAS is superior in nearly all cases; the only exception is ReasonVOS with InternVL3-8B (57.2 vs.\ 58.9), likely due to dynamic tiling weakening correspondence with our fixed spatial grid. ATLAS trains only a lightweight attention adapter while keeping both the LVLM and SAM2 frozen, using only the Ref-YouTube-VOS~\cite{Refer-Youtube-VOS} training split for about six GPU hours, yet generalizes effectively to unseen reasoning video segmentation benchmarks.

\subsection{Evaluation on Additional Datasets}
To further evaluate the generalization of ATLAS, we report results on the MeViS \cite{Mevis} and Ref-YouTube-VOS \cite{Refer-Youtube-VOS} benchmarks in Table~\ref{tab:additional-datasets}. ATLAS consistently achieves competitive performance across both datasets.

On MeViS, ATLAS with Qwen2.5VL-7B attains a $\mathcal{J}\&\mathcal{F}$ score of 54.5, outperforming both DecAF (48.1) and the best fully fine-tuned method, Veason-R1 (52.2). Similar improvements are observed with the other LVLM backbones, demonstrating that the learned attention adaptation generalizes well across different visual-language models.

On Ref-YouTube-VOS, ATLAS achieves competitive performance. In particular, using Qwen2VL-7B, it obtains a $\mathcal{J}\&\mathcal{F}$ score of 71.0, substantially outperforming previous frozen-backbone methods and approaching the performance of the best fully fine-tuned approaches. These results further demonstrate that adapting intermediate LVLM attention representations provides an effective and computationally efficient alternative to end-to-end fine-tuning.

\begin{table*}[!t]
\centering
\vspace{0.3cm}
\begin{minipage}[t]{0.6\textwidth}
\centering
\tiny
\caption{Results on MeViS~\cite{Mevis} and Ref-YouTube-VOS~\cite{Refer-Youtube-VOS}. $^\dagger$ATLAS is trained on Ref-YouTube-VOS train set and evaluated on val set; all others are zero-shot.}

\label{tab:additional-datasets}
\definecolor{baselineorange}{RGB}{242,247,255}

\resizebox{\linewidth}{!}{%
\begin{tabular}{llcccccc}
\toprule
\multirow{2}{*}{Method} & \multirow{2}{*}{LVLM} & \multicolumn{3}{c}{MeViS} & \multicolumn{3}{c}{Ref-YouTube-VOS$^\dagger$} \\
\cmidrule(lr){3-5} \cmidrule(lr){6-8}
& & $\mathcal{J\&F}$ & $\mathcal{J}$ & $\mathcal{F}$ & $\mathcal{J\&F}$ & $\mathcal{J}$ & $\mathcal{F}$ \\
\midrule

\rowcolor{baselineorange}
VISA \cite{VISA}        & Chat-UniVi-7B  & 44.5 & 41.8 & 47.1 & 61.5 & {59.8} & 63.2 \\
\rowcolor{baselineorange}
VideoLISA \cite{reasonvos-videolisa} & LLaVA-Phi-3-V  & 44.4 & 41.3 & 47.6 & 63.7 & 61.7 & 65.7 \\
\rowcolor{baselineorange}
GLUS \cite{GLUS}       & LLaVA-7B       & 51.3 & {48.5} & 54.2 & 67.3 & 65.5 & 69.0 \\
\rowcolor{baselineorange}
VRS-HQ \cite{VRS-HQ}      & Chat-UniVi-7B  & 50.6 & 47.6 & 53.7 & {70.4} & {68.3} & \textbf{72.5} \\
\rowcolor{baselineorange}
Veason-R1 \cite{Veason-R1}& Qwen2.5VL-7B   & {52.2} & 48.4 & {56.0} & -- & -- & -- \\

\midrule

\rowcolor{baselinegreen}
Loc-Head \cite{LOC-HEAD} & Qwen2.5VL-7B & 39.4 & 35.5 & 43.6 & 51.0 & 46.8 & 55.2 \\
\rowcolor{baselinegreen}
DecAF \cite{DecAF} & Qwen2.5VL-7B & 48.1 & 44.0 & 52.1 & 59.9 & 56.2 & 63.5 \\
\midrule
\rowcolor{greenrow}
{ATLAS} & Qwen2.5VL-7B & \textbf{54.5} & \textbf{51.0} & \textbf{58.0} & 69.5 & 67.7 & 71.3 \\
\rowcolor{greenrow}
{ATLAS} & Qwen2VL-7B & {52.1} & {48.4} & {55.8} & \textbf{71.0} & \textbf{69.2} & \textbf{72.8} \\
\rowcolor{greenrow}
{ATLAS} & LLaVA-OV-7B & 46.0 & 42.2 & 49.9 & 59.9 & 58.1 & 61.7 \\
\rowcolor{greenrow}
{ATLAS} & InternVL3-8B & {51.4} & 47.9 & 54.8 & 66.0 & 64.2 & 67.8 \\

\bottomrule
\end{tabular}
}

\end{minipage}
\hspace{0.01\textwidth}
\vrule
\hspace{0.01\textwidth}
\begin{minipage}[t]{0.36\textwidth}
\centering

\scriptsize
\setlength{\tabcolsep}{9pt}
\renewcommand{\arraystretch}{1}

\captionof{table}{Localization quality of adapted attention maps on sampled keyframes.}
\label{tab:localization_quality}

\begin{tabular}{@{}llccc@{}}
\toprule
\textbf{Dataset} & \textbf{Method} & $\mathbf{\mathcal{J}}$ & $\mathbf{\mathcal{F}}$ & $\mathbf{\mathcal{J}\&\mathcal{F}}$ \\
\midrule
\multirow{2}{*}{Ref-DAVIS}
& ATLAS & \textbf{47.2} & \textbf{74.0} & \textbf{60.6} \\
& Raw Attn. & 2.0 & 18.2 & 10.1 \\
\addlinespace
\multirow{2}{*}{ReasonVOS}
& ATLAS & \textbf{40.3} & \textbf{57.4} & \textbf{48.9} \\
& Raw Attn. & 4.3 & 27.3 & 15.8 \\
\bottomrule
\end{tabular}

\vspace{0.3cm}

\includegraphics[width=\linewidth]{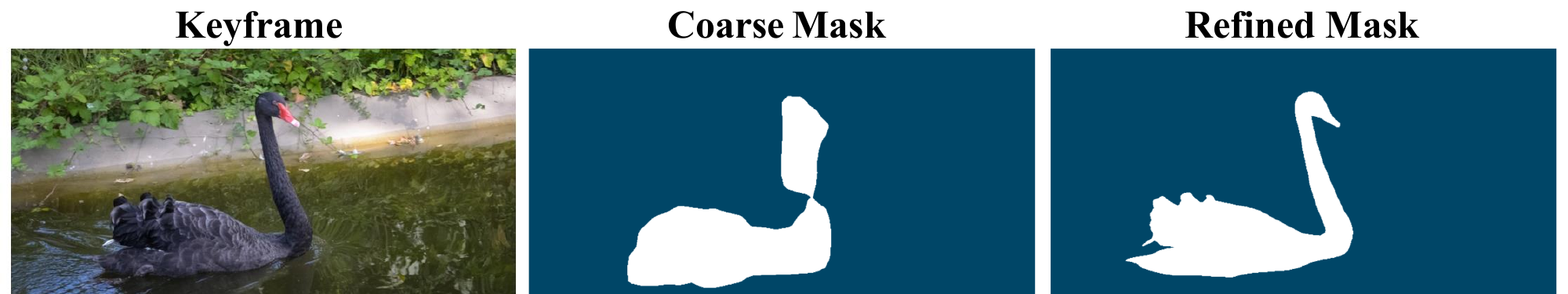}

\captionof{figure}{Examples of thresholded localization maps predicted by ATLAS.}
\label{fig:localization_examples}

\end{minipage}

 \end{table*}

\begin{figure*}[!t]
    \centering
    \includegraphics[width=\linewidth]{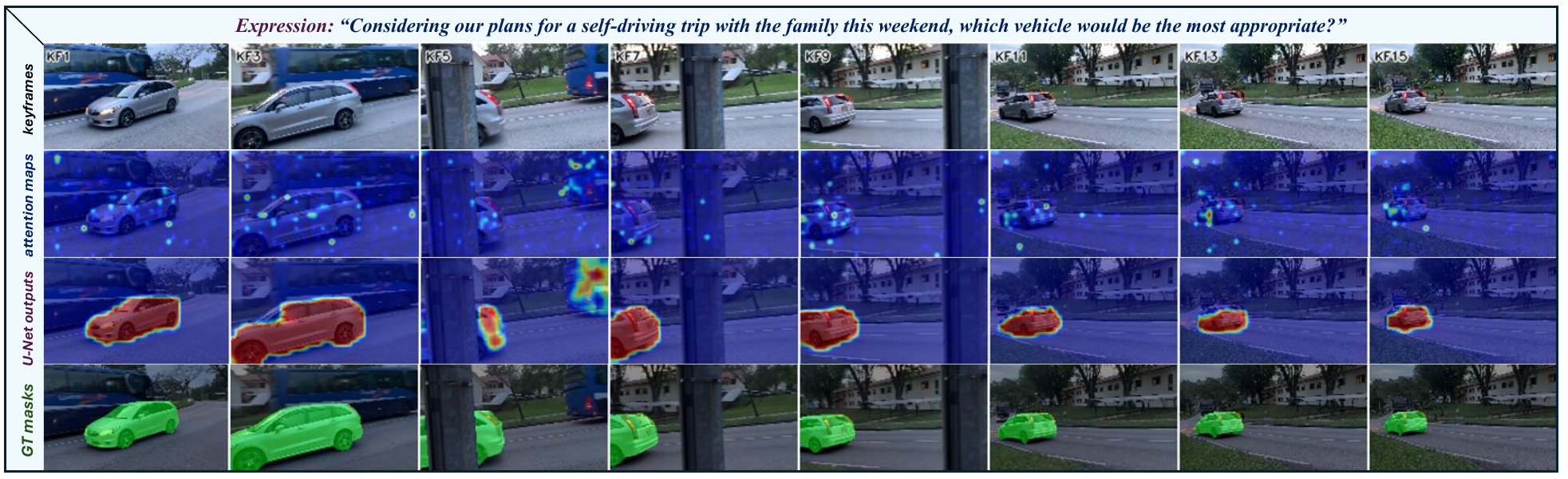}
    \caption{Attention adaptation for reasoning video segmentation. From top to bottom: sampled keyframes, averaged LVLM attention maps, U-Net-adapted coarse attention maps, and ground-truth masks. }
    \label{fig:qualitative_example}
\end{figure*}

\subsection{Attention Localization Quality}

We evaluate the attention adapter independently of SAM2 by thresholding its sigmoid output at 0.5 on sampled keyframes and comparing the resulting masks with ground truth. As a baseline, we apply the same threshold to normalized, averaged raw LVLM frame-level attentions. No propagation or post-processing is used.

As shown in Table~\ref{tab:localization_quality}, adaptation substantially improves localization, raising $\mathcal{J}\&\mathcal{F}$ from 10.1 to 60.6 on Ref-DAVIS and from 15.8 to 48.9 on ReasonVOS. Figure~\ref{fig:localization_examples} shows that the adapted map provides accurate coarse localization for guiding SAM2, confirming that ATLAS transforms diffuse LVLM attentions into reliable cues for mask propagation.

\subsection{Qualitative Analysis}

Figure~\ref{fig:qualitative_example} compares raw averaged attentions, ATLAS-adapted maps, and ground-truth masks. While raw attentions are often diffuse and affected by attention sinks, ATLAS produces object-centric localization maps that suppress irrelevant regions and reliably guide SAM2 toward temporally consistent masks. Additional examples, training dynamics, and intermediate localization maps are provided in Appendix~\ref{app:additional-examples} and Appendix~\ref{app:training-dynamics}, illustrating how adaptation progressively converts noisy LVLM attentions into accurate object localization cues.

\begin{table}[t]
\centering
\vspace{-0.2cm}
\caption{Temporal and video ablation with Qwen2.5-VL-7B.}
\label{tab:temp_video_2x2}

\resizebox{\columnwidth}{!}{%
\begin{tabular}{lccc ccc}
\toprule
& \multicolumn{3}{c}{Ref-DAVIS} & \multicolumn{3}{c}{ReasonVOS} \\
\cmidrule(lr){2-4}\cmidrule(lr){5-7}
Method & $\mathcal{J}$ & $\mathcal{F}$ & $\mathcal{J\&F}$
       & $\mathcal{J}$ & $\mathcal{F}$ & $\mathcal{J\&F}$ \\
\midrule
 Frame only      & 75.0 & 80.1 & 77.5 & 57.8 & 63.0 & 60.4 \\

 Frame + Video    & 76.3 & 82.0 & 79.1 & 62.2 & 67.7 & 64.9 \\

 Frame + FiLM   & 75.5 & 81.3 & 78.4 & 58.8 & 64.1 & 61.5 \\

\rowcolor{baselinegreen} Frame + Video + FiLM & \bf 76.8 & \bf 83.4 & \bf 80.1 & \bf 64.7 & \bf 70.2 & \bf 67.4 \\
\bottomrule
\end{tabular}%
}
\end{table}
\section{Ablation Study}
\label{sec:ablation}

\subsection{Effect of Video Attention and Temporal Modulation}
\label{sec:ablation_temporal}

We jointly ablate ATLAS's two video-specific components: the video-attention branch and temporal FiLM modulation. Removing both yields a per-frame baseline using only frame-level attention, approximating the image-domain F-LMM design~\cite{F-LMM} with frozen-SAM2 propagation. Table~\ref{tab:temp_video_2x2} shows that the components are complementary. On ReasonVOS, video attention alone improves $\mathcal{J}\&\mathcal{F}$ by $+4.5$ points (60.4$\rightarrow$64.9), temporal FiLM by $+1.1$ (60.4$\rightarrow$61.5), and their combination by $+7.0$ (60.4$\rightarrow$67.4). On Ref-DAVIS, the corresponding gains are $+1.6$, $+0.9$, and $+2.6$ points. The video branch provides cross-frame context that FiLM injects into each keyframe's spatial decoding. The larger gain on ReasonVOS than Ref-DAVIS ($+7.0$ vs.\ $+2.6$ in $\mathcal{J}\&\mathcal{F}$) suggests that reasoning expressions rely more strongly on temporal cues, requiring the model to capture both \emph{when} the object is relevant and \emph{where} it is located.

\begin{table}[t]
\centering
\definecolor{baselineorange}{RGB}{242,247,255}
\caption{Tracklet selection criterion using Qwen2.5-VL-7B. All variants share identical localization maps and SAM2 propagation.}
\vspace{-0.2cm}
\label{tab:selection}
\scriptsize
\setlength{\tabcolsep}{6pt}
\begin{tabular}{lccc|ccc}
\toprule
& \multicolumn{3}{c|}{Ref-DAVIS}
& \multicolumn{3}{c}{ReasonVOS} \\
Criterion & $\mathcal{J}$ & $\mathcal{F}$ & $\mathcal{J}\&\mathcal{F}$
          & $\mathcal{J}$ & $\mathcal{F}$ & $\mathcal{J}\&\mathcal{F}$ \\
\midrule
\rowcolor{baselineorange}Random             & 37.1 & 45.8 & 41.4 & 27.4 & 34.2 & 30.8 \\
SAM2 score         & 51.5 & 59.8 & 55.6 & 37.8 & 44.3 & 41.1 \\
Mean prob.         & 52.8 & 59.2 & 56.0 & 48.0 & 53.9 & 50.9 \\
Captured mass      & 70.2 & 77.0 & 73.6 & 48.4 & 54.8 & 51.6 \\
Hard IoU           & 75.0 & 81.2 & 78.1 & 63.4 & 68.8 & 66.1 \\
Soft IoU    & 75.6 & 81.1 & 78.3 & 59.7 & 65.4 & 62.6 \\
 Cosine Similarity           & 75.2 & 81.1 & 78.1 & 60.7 & 66.4 & 63.6 \\
Volumetric Pearson & 76.7 & 83.1 & 79.9 & 61.5 & 66.9 & 64.2 \\
\rowcolor{baselinegreen} \textbf{Per-frame Pearson} & \textbf{76.8} & \textbf{83.4} & \textbf{80.1} & \textbf{64.7} & \textbf{70.2} & \textbf{67.4} \\

\midrule
\rowcolor{baselineorange}\textit{Oracle}    & \textit{83.1} & \textit{89.7} & \textit{86.4} & \textit{79.6} & \textit{85.6} & \textit{82.6} \\
\bottomrule
\end{tabular}
\vspace{-0.2cm}
\end{table}

\begin{table}[t]
\tiny
\centering
\caption{Effect of the number of keyframes on ReasonVOS.}
\vspace{-0.2cm}
\label{tab:keyframe_ablation}
\renewcommand{\arraystretch}{0.9}
\resizebox{\columnwidth}{!}{
\begin{tabular}{c cccccc}
\toprule
\# Keyframes 
 & 4 & 8 & 12 & 16 & 20 & 24 \\
\midrule
$\mathcal{J}$ 
& 59.4 & 61.6 & 63.0 & \cellcolor{baselinegreen}64.7 & 63.5 & 64.2 \\

$\mathcal{F}$ 
& 64.5 & 66.8 & 68.9 & \cellcolor{baselinegreen}70.2 & 70.1 & 70.4 \\

$\mathcal{J}\&\mathcal{F}$ 
& 62.0 & 64.2 & 65.9 & \cellcolor{baselinegreen}67.4 & 66.8 & 67.3 \\
\bottomrule
\end{tabular}
}
\end{table}

\subsection{Effect of the Tracklet Selection Criterion}
\label{sec:ablation_selection}
We study the similarity criterion \(\rho(\cdot,\cdot)\) used for candidate selection in Eq.~\ref{eq:eq12}, comparing per-frame and volumetric Pearson correlation with SAM2 confidence, mean probability, captured mass, hard IoU, soft IoU, cosine similarity, random selection, and a ground-truth oracle. Table~\ref{tab:selection} shows that per-frame Pearson performs best among practical criteria on both benchmarks, outperforming volumetric Pearson and confidence-, mass-, and overlap-based alternatives. Its scale- and offset-invariance suits diffuse, uncalibrated LVLM attentions, while frame-wise comparison preserves temporal variation. The oracle reaches 86.4 and 82.6 \(\mathcal{J}\&\mathcal{F}\), indicating room for improved selection. We therefore use per-frame Pearson correlation as \(\rho\) throughout.

\subsection{Effect of Keyframe Number}

Table~\ref{tab:keyframe_ablation} studies the effect of the number of sampled keyframes on ReasonVOS. Performance generally improves as more keyframes are used, with $\mathcal{J}\&\mathcal{F}$ increasing from 62.0 at 4 keyframes to a peak of 67.4 at 16. Beyond 16 keyframes, the gains saturate and slightly fluctuate, suggesting that additional frames provide limited new temporal information while increasing computation. 

\subsection{Effect of the Adapter Architecture}
\label{sec:ablation_adapter}

We compare the U-Net with simpler supervised heads using the same frame+video attention input, training schedule, frozen SAM2, and tracklet selection: a linear probe, a two-layer MLP, and a single-scale convolutional head. Table~\ref{tab:adapter} shows that the linear probe performs poorly in $\mathcal{J}\&\mathcal{F}$ ($57.5$ on Ref-DAVIS and $48.3$ on ReasonVOS), while the MLP and convolutional head improve substantially but remain $4$--$5$ points below the U-Net. Their similar performance ($73.2$ vs.\ $74.3$ on Ref-DAVIS and $61.9$ vs.\ $61.3$ on ReasonVOS) suggests that local spatial context alone is insufficient, and that the U-Net gains mainly come from multi-scale aggregation. Temporal modulation (FiLM) further improves $\mathcal{J}\&\mathcal{F}$ by $0.8$ points on Ref-DAVIS and $2.5$ points on ReasonVOS.

\begin{table}[t]
\centering
\caption{Adapter architecture ablation with Qwen2.5-VL-7B.}
\vspace{-0.2cm}
\label{tab:adapter}
\tiny
\setlength{\tabcolsep}{4pt}
\resizebox{\columnwidth}{!}{%
\begin{tabular}{lccc|ccc}
\toprule
& \multicolumn{3}{c|}{Ref-DAVIS} 
& \multicolumn{3}{c}{ReasonVOS} \\
Adapter 
& $\mathcal{J}$ 
& $\mathcal{F}$ 
& $\mathcal{J}\&\mathcal{F}$
& $\mathcal{J}$ 
& $\mathcal{F}$ 
& $\mathcal{J}\&\mathcal{F}$ \\
\midrule
Linear       & 53.6 & 61.5 & 57.5 & 45.3 & 51.3 & 48.3 \\
MLP          & 70.1 & 76.2 & 73.2 & 59.4 & 64.5 & 61.9 \\
Conv3        & 71.2 & 77.3 & 74.3 & 58.6 & 63.9 & 61.3 \\
\rowcolor{baselinegreen} U-Net        & 76.3 & 82.0 & 79.1 & 62.2 & 67.7 & 64.9 \\
\midrule
\rowcolor{greenrow}
U-Net + FiLM & \textbf{76.8} & \textbf{83.4} & \textbf{80.1}
             & \textbf{64.7} & \textbf{70.2} & \textbf{67.4} \\
\bottomrule
\end{tabular}%
}
\end{table}

\subsection{Runtime Analysis}
\label{sec:runtime}

Since DecAF~\cite{DecAF} couples its budgets ($\text{frame}=4\times\text{video}$), we compare at the matched 256-video-token setting and also report ATLAS at $64/1024$, used elsewhere, with Qwen2.5-VL-7B. Table~\ref{tab:runtime} shows that ATLAS achieves a $2.7\times$ end-to-end speedup ($15.6$s vs.\ $41.3$s) while improving $\mathcal{J}\&\mathcal{F}$ ($67.5$ vs.\ $63.9$). The speedup comes from faster extraction ($2.46$s vs.\ $8.08$s), using a single generated token instead of repeated decoding and contrastive passes, and faster propagation ($13.1$s vs.\ $33.2$s) via single-point seeding and duplicate-track suppression. Reducing the video branch from 256 to 64 tokens costs only $0.1$ $\mathcal{J}\&\mathcal{F}$, suggesting it mainly provides complementary temporal cues.

\begin{table}[t]
\centering
\caption{Runtime and accuracy on an NVIDIA RTX PRO 6000.}
\vspace{-0.2cm}
\label{tab:runtime}
\small
\setlength{\tabcolsep}{3pt}
\resizebox{\columnwidth}{!}{%
\begin{tabular}{llcccccc}
\toprule
Method & vid/frm & Attn. & Adapt. & Seed & SAM2 & E2E & $\mathcal{J}\&\mathcal{F}$ \\
\midrule
ATLAS (ours) & 64\,/\,1024  & \bf 2.06 & 0.003 & 0.001 & \bf 13.2 & \textbf{15.3} & \bf 67.4 \\
\midrule
ATLAS & 256\,/\,1024 & \cellcolor{baselinegreen} \bf 2.46 & 0.003 & 0.001 &  \cellcolor{baselinegreen} \bf 13.1 &  \cellcolor{baselinegreen} \textbf{15.6} & \bf 67.5 \\
DecAF~\cite{DecAF} & 256\,/\,1024 &  \cellcolor{baselinegreen} 8.08 & -- & 0.030 &  \cellcolor{baselinegreen} 33.2 &  \cellcolor{baselinegreen} 41.3 & 63.9 \\
\bottomrule
\end{tabular}%
}
\vspace{-0.3cm}
\end{table}

\vspace{-0.1cm}

\section{Conclusion and Limitations}

We presented ATLAS, an attention adaptation framework for reasoning video object segmentation that transforms diffuse LVLM attentions across layers, heads, modalities, and time into object-centric localization maps that guide a frozen SAM2 model. Without fine-tuning either backbone, ATLAS outperforms frozen-backbone methods and remains competitive with fully fine-tuned approaches, showing that LVLM attentions contain rich localization cues that can be unlocked through learned adaptation. ATLAS, however, adapts attention rather than correcting LVLM reasoning, so severe reasoning failures may still limit localization. It also assumes access to intermediate visual attentions and uses uniform keyframe sampling. Future work could improve reasoning, support models with restricted attention access, and explore adaptive keyframe selection for long or dynamic videos.

{
    \small
    \bibliographystyle{ieeenat_fullname}
    \bibliography{main}
}
\clearpage

\appendix

\section{Object Extraction Prompt}
\label{app:prompt}

\begin{center}
\vspace{0.3cm}
\begin{tcolorbox}[
    enhanced,
    width=\linewidth,
    colback=gray!5,
    colframe=black!20,
    boxrule=0.5pt,
    arc=2mm,
    left=3mm,
    right=3mm,
    top=2mm,
    bottom=2mm,
    title=\textbf{Object Extraction Prompt},
    fonttitle=\bfseries,
    coltitle=black,
]

\textbf{Input Expression}

\vspace{0.4em}

\begin{center}
\fbox{%
\parbox{0.85\textwidth}{%
\centering
\itshape
\{Expression\}
}}
\end{center}

\vspace{0.8em}

\textbf{Prompt}

\begin{quote}\small
What is the main object (or objects) referred to in the given expression or question?

Focus on the \textbf{primary subject or agent} involved in the described action or behavior.

Respond with a \textbf{single word} (e.g., ``cat'', ``person'', ``dog'') that best describes the target object(s).
\end{quote}

\end{tcolorbox}

\captionof{figure}{Prompt used to extract the target object from the referring expression.}
\label{fig:object_prompt}
\end{center}

To obtain informative visual attention maps from the frozen LVLM, we first augment the referring expression with a fixed object extraction prompt. Rather than directly prompting the model to localize the referred object, we ask it to identify the primary object category mentioned in the expression and generate a single response token (Figure \ref{fig:object_prompt}). We then extract the attention of this generated token to all visual tokens across transformer layers and attention heads.

\section{More Implementation Details}
\label{sec:appendix-imlementation-details}
The training hyperparameters of the proposed attention adapter are summarized in Table~\ref{tab:atlas_train}. We optimize the network using AdamW with an initial learning rate of $10^{-4}$ that linearly decays to $10^{-5}$ over 20 epochs. A weight decay of $10^{-4}$ is applied for regularization, and training is performed with a batch size of eight video clips. The network is supervised using a combination of binary cross-entropy with logits and Soft Dice loss, which jointly encourage accurate pixel-wise classification and robust mask overlap.

Table~\ref{tab:atlas_arch} summarizes the architecture of the proposed attention adaptation network. The network follows a U-Net-style encoder--decoder design operating on concatenated frame- and video-level LVLM attention features. Given $C{=}392$ channels from each modality, the input contains $2C{=}784$ channels over a $24{\times}42$ spatial grid. The first encoder stage applies two $3{\times}3$ convolutions to produce 128 channels, followed by $2{\times}2$ max pooling. The second encoder stage increases the feature dimension to 256 channels at half the spatial resolution.

At the bottleneck, two additional $3{\times}3$ convolutions produce a 512-channel representation. To capture temporal dependencies across keyframes, the bottleneck feature of each frame is globally average pooled and passed through a two-layer, four-head temporal transformer. Its output is mapped to FiLM parameters $\gamma$ and $\beta$, which modulate the 512-channel bottleneck features channel-wise. This conditions the spatial decoding of each frame on information from the full video sequence rather than processing frames independently.

The decoder then progressively restores the spatial resolution using transposed convolutions and U-Net skip connections. The first upsampling stage fuses the temporally modulated bottleneck with Encoder-2 features and reduces the representation to 256 channels. The second stage combines it with Encoder-1 features and restores the original $R_f{\times}W_f$ resolution with 128 channels. Finally, a $1{\times}1$ convolution produces a single-channel localization logit map, which provides object-centric spatial cues for the frozen SAM2 model.

\begin{table}[!t]
\centering
\caption{Training hyperparameters of the ATLAS adapter.}
\label{tab:atlas_train}
\small
\setlength{\tabcolsep}{6pt}
\renewcommand{\arraystretch}{1.15}
\begin{tabular}{ll}
\toprule
\textbf{Hyperparameter} & \textbf{Value} \\
\midrule
Optimizer & AdamW \\
Learning rate & $10^{-4}\rightarrow10^{-5}$ (linear decay) \\
Weight decay & $10^{-4}$ \\
Epochs & 20 \\
Batch size & 8 clips \\
Segmentation loss & BCE-with-Logits + Soft Dice \\
\bottomrule
\end{tabular}
\end{table}

\begin{table*}[!t]
\centering
\caption{ATLAS Temporal FiLM FV U-Net architecture ($C{=}392$, grid $R_f{\times}W_f{=}24{\times}42$).}
\label{tab:atlas_arch}
\small
\setlength{\tabcolsep}{5pt}
\renewcommand{\arraystretch}{1.15}
\begin{tabular}{lccc}
\toprule
\textbf{Stage} & \textbf{Operation} & \textbf{Output} & \textbf{Spec.} \\
\midrule
Input
& Frame / Frame+Video Attn.
& $2C \times R_f \times W_f$
& $784 \times 24 \times 42$ \\

Encoder-1
& $3{\times}3$ Conv $\times 2$ 
& $128 \times R_f \times W_f$
& BN + ReLU \\

Downsample
& $2{\times}2$ MaxPool
& $128 \times {R_f}/{2} \times {W_f}/{2}$
& -- \\

Encoder-2
& $3{\times}3$ Conv $\times 2$ 
& $256 \times {R_f}/{2} \times {W_f}/{2}$
& BN + ReLU \\

Bottleneck
& $3{\times}3$ Conv $\times 2$ 
& $512 \times {R_f}/{2} \times {W_f}/{2}$
& -- \\

Temporal
& GAP $\rightarrow$ Transformer
& $512$
& 2 layers, 4 heads \\

FiLM
& $\gamma, \beta$
& $512$
& Channel-wise modulation \\

Upsample-2
& Transposed Conv + skip (enc2)
& $256 \times {R_f}/{2} \times {W_f}/{2}$
& Concat skip \\

Decoder-2
& $3{\times}3$ Conv $\times 2$ 
& $256 \times {R_f}/{2} \times {W_f}/{2}$
& BN + ReLU \\

Upsample-1
& Transposed Conv + skip (enc1)
& $128 \times R_f \times W_f$
& Concat skip \\

Decoder-1
& $3{\times}3$ Conv $\times 2$ 
& $128 \times R_f \times W_f$
& BN + ReLU \\

Output
& $1{\times}1$ Conv
& $1 \times R_f \times W_f$
& Segmentation logits \\
\bottomrule
\end{tabular}
\end{table*}

\begin{figure*}[!t]
    \centering
    \includegraphics[width=\linewidth]{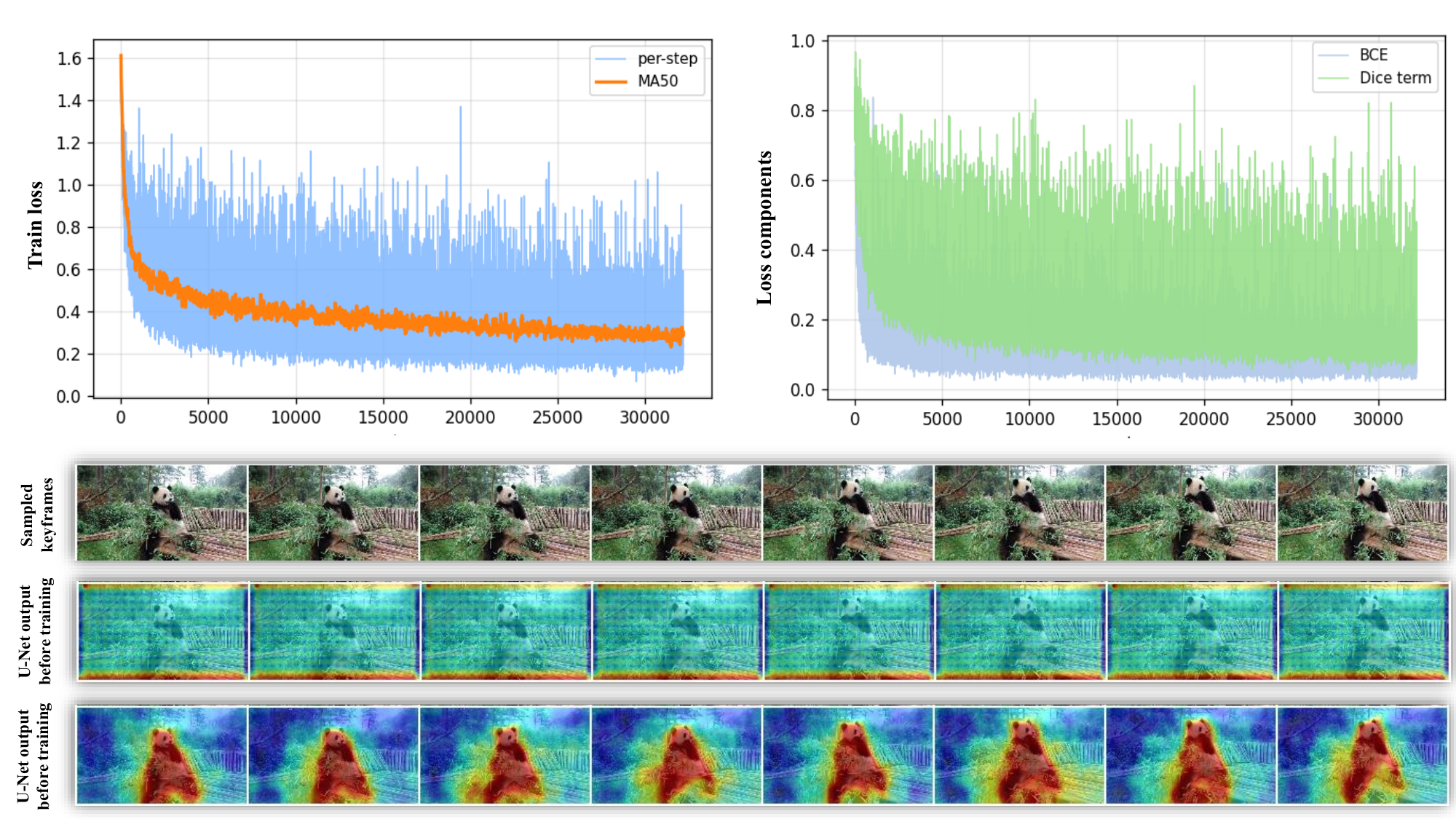}
    \caption{Training convergence of the U-Net adapter and qualitative outputs before and after training.}
    \label{fig:training_dynamics}
\end{figure*}

\section{Training Dynamics}
Figure~\ref{fig:training_dynamics} shows stable adapter training, with the total, binary cross-entropy, and Soft Dice losses decreasing consistently. Qualitatively, the adapter progressively transforms diffuse LVLM attentions into object-centric localization maps by suppressing background responses and concentrating on the referred object, producing accurate cues for the frozen SAM2 model.
\label{app:training-dynamics}

\section{Additional Qualitative Results}
\label{app:additional-examples}

\begin{figure*}
    \centering
    \includegraphics[width=\linewidth]{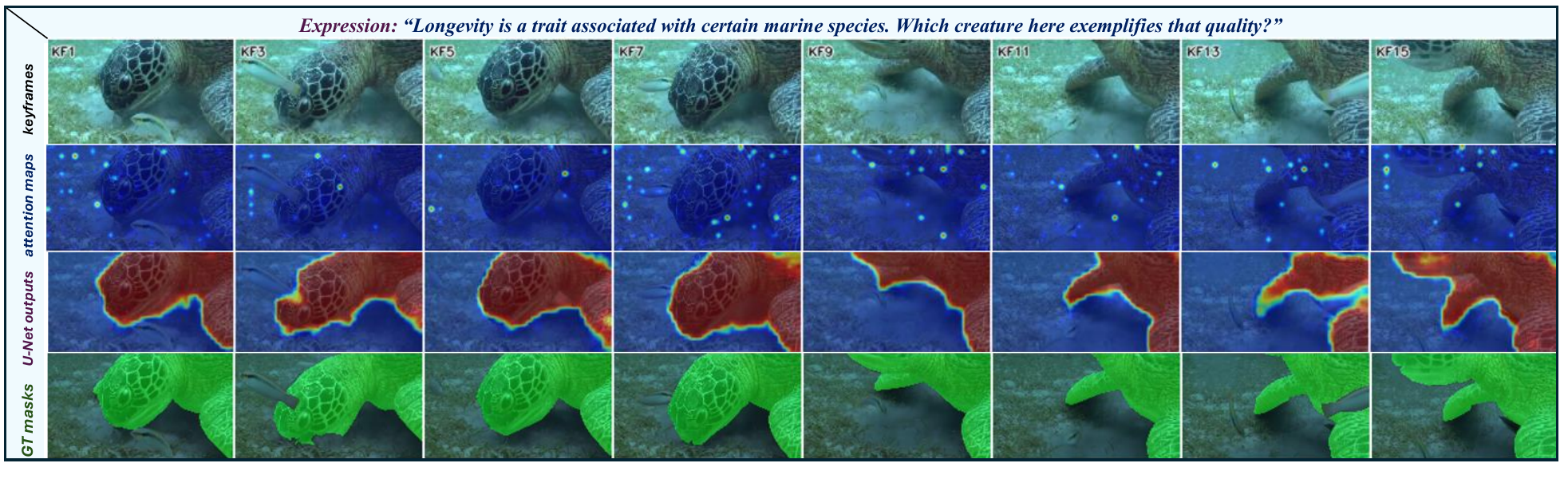}
    \caption{Additional qualitative example. From top to bottom: input frames, raw averaged attention maps, adapted attention maps produced by ATLAS, and the corresponding SAM2 segmentation masks.}
    \label{fig:7}
\end{figure*}

\begin{figure*}
    \centering
    \includegraphics[width=\linewidth]{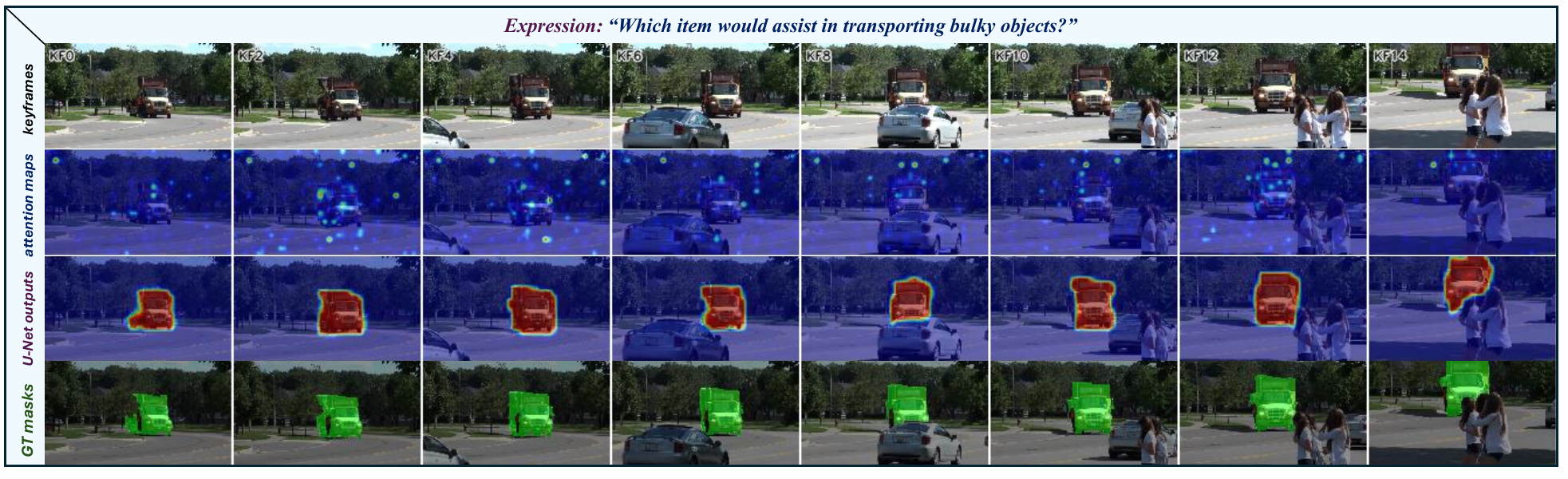}
       \caption{Additional qualitative example. From top to bottom: input frames, raw averaged attention maps, adapted attention maps produced by ATLAS, and the corresponding SAM2 segmentation masks.}
    \label{fig:8}
\end{figure*}

\begin{figure*}
    \centering
    \includegraphics[width=\linewidth]{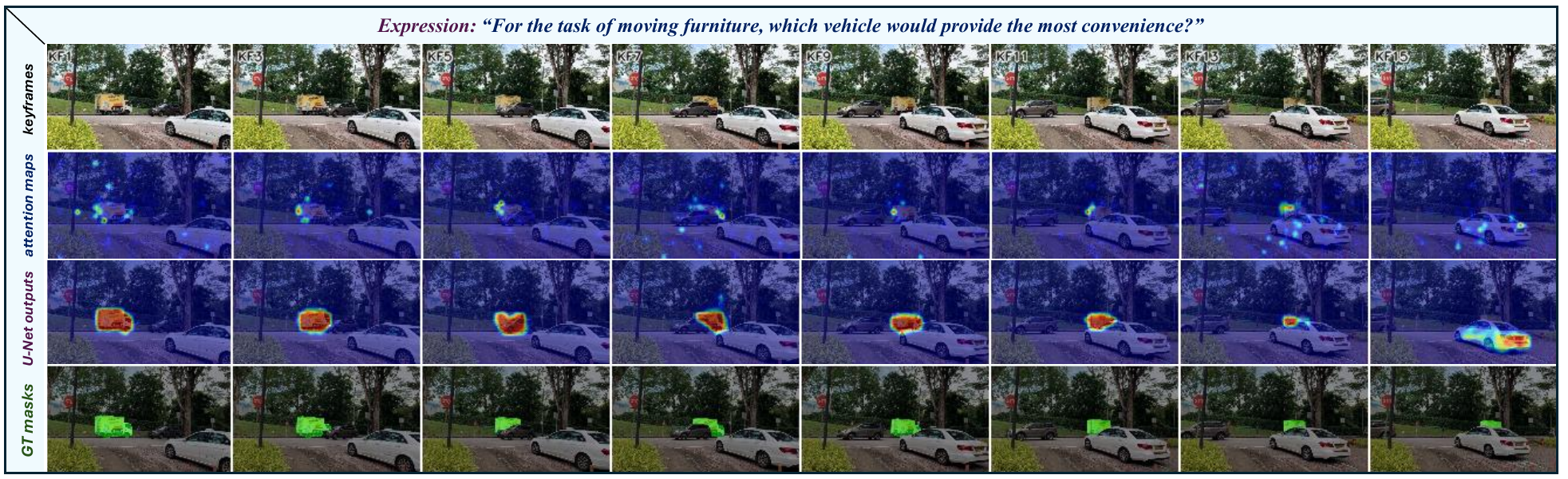}
     \caption{Additional qualitative example. From top to bottom: input frames, raw averaged attention maps, adapted attention maps produced by ATLAS, and the corresponding SAM2 segmentation masks.}
    \label{fig:9}
\end{figure*}

\begin{figure*}
    \centering
    \includegraphics[width=\linewidth]{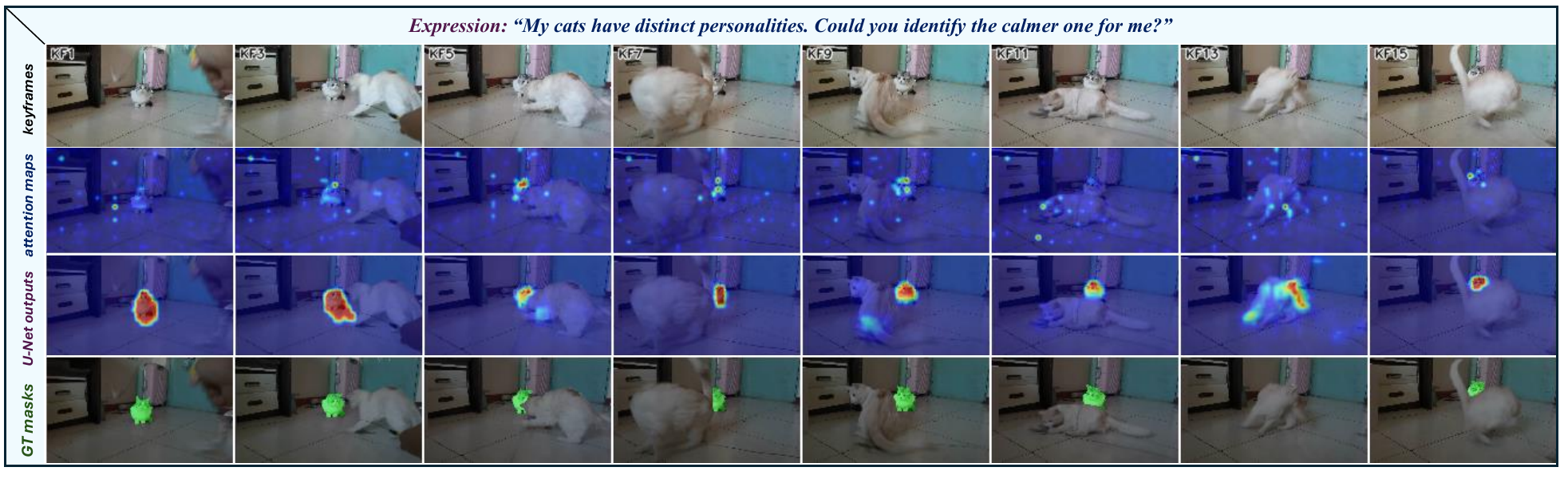}
        \caption{Additional qualitative example. From top to bottom: input frames, raw averaged attention maps, adapted attention maps produced by ATLAS, and the corresponding SAM2 segmentation masks.}
    \label{fig:10}
\end{figure*}

\begin{figure*}
    \centering
    \includegraphics[width=\linewidth]{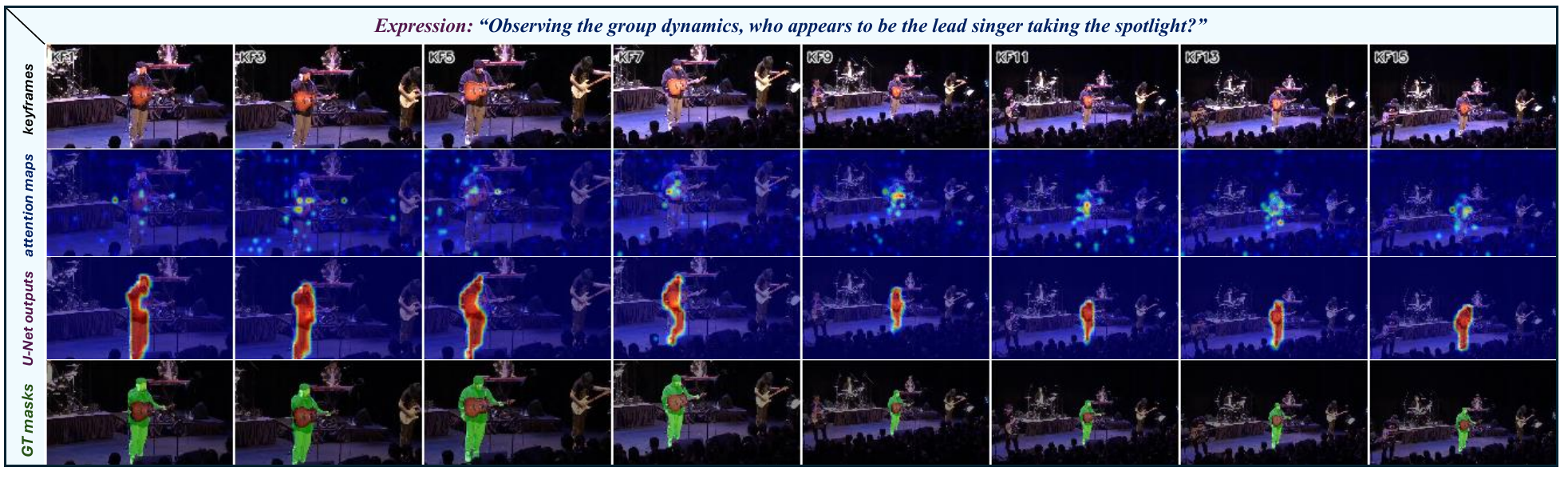}
        \caption{Additional qualitative example. From top to bottom: input frames, raw averaged attention maps, adapted attention maps produced by ATLAS, and the corresponding SAM2 segmentation masks.}
    \label{fig:11}
\end{figure*}

\begin{figure*}
    \centering
    \includegraphics[width=\linewidth]{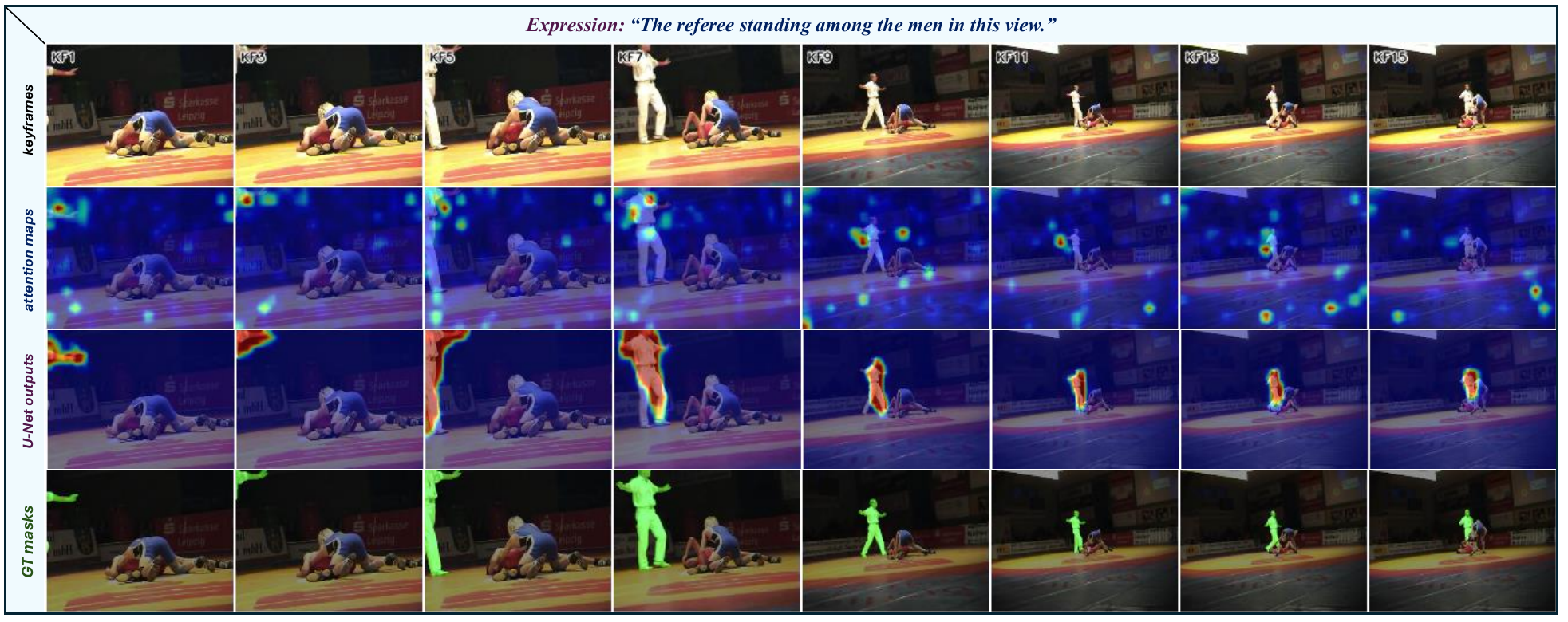}
       \caption{Additional qualitative example. From top to bottom: input frames, raw averaged attention maps, adapted attention maps produced by ATLAS, and the corresponding SAM2 segmentation masks.}
    \label{fig:12}
\end{figure*}

\begin{figure*}
    \centering
    \includegraphics[width=\linewidth]{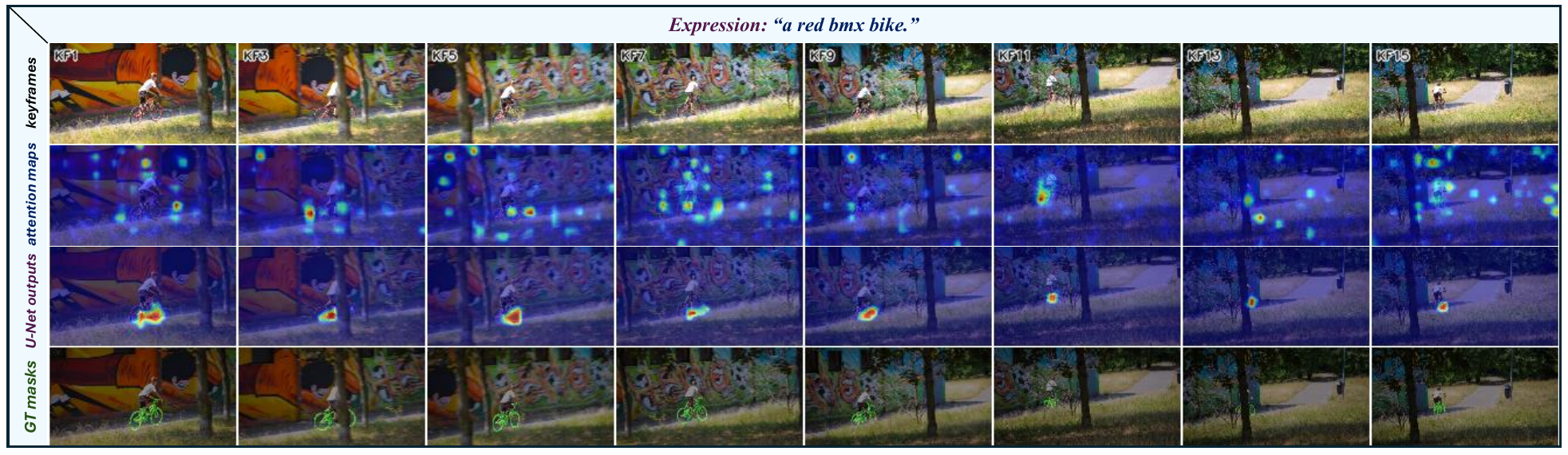}
        \caption{Additional qualitative example. From top to bottom: input frames, raw averaged attention maps, adapted attention maps produced by ATLAS, and the corresponding SAM2 segmentation masks.}
    \label{fig:13}
\end{figure*}

\begin{figure*}
    \centering
    \includegraphics[width=\linewidth]{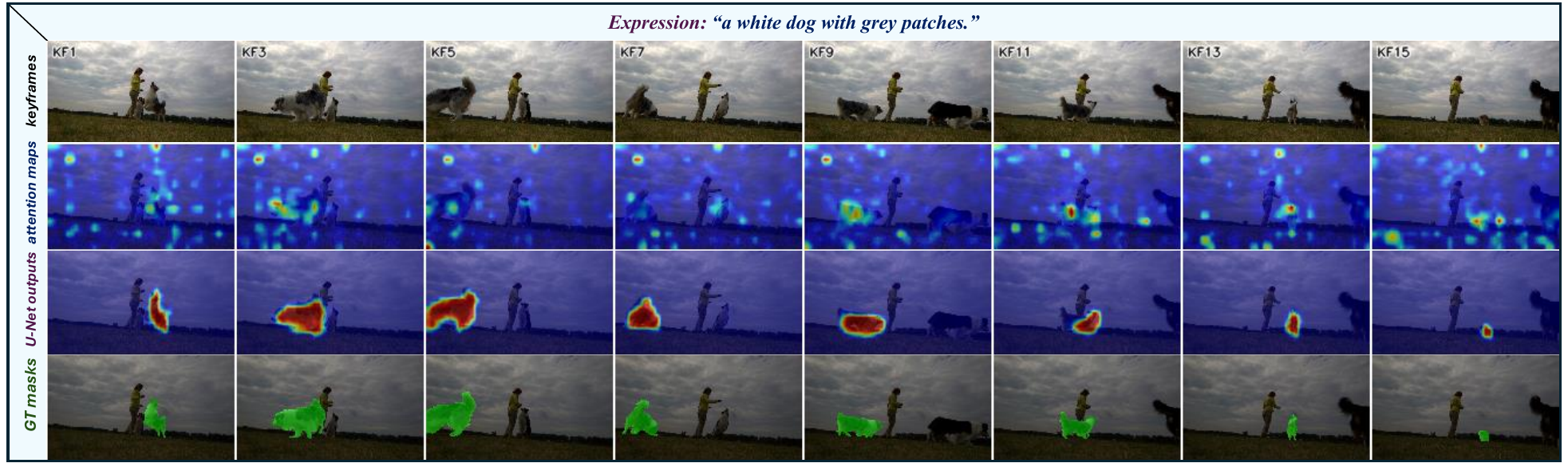}
        \caption{Additional qualitative example. From top to bottom: input frames, raw averaged attention maps, adapted attention maps produced by ATLAS, and the corresponding SAM2 segmentation masks.}
    \label{fig:14}
\end{figure*}

\begin{figure*}
    \centering
    \includegraphics[width=\linewidth]{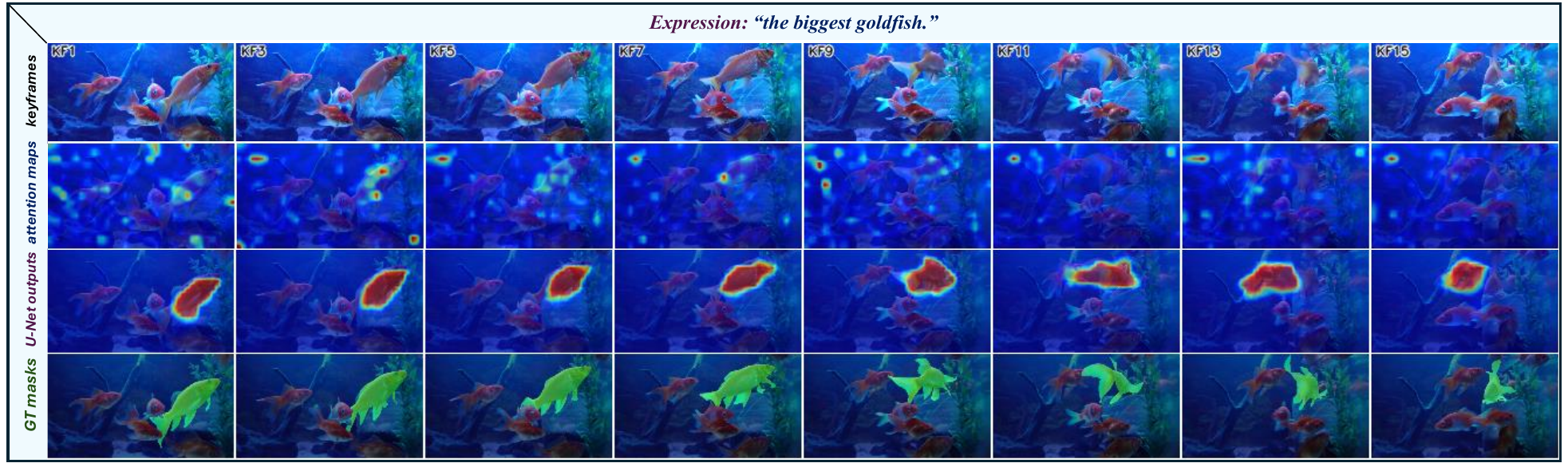}
        \caption{Additional qualitative example. From top to bottom: input frames, raw averaged attention maps, adapted attention maps produced by ATLAS, and the corresponding SAM2 segmentation masks.}
    \label{fig:15}
\end{figure*}

Figure~\ref{fig:7}-\ref{fig:15} provide additional qualitative examples, demonstrating that ATLAS consistently produces object-centric attention maps that guide accurate segmentation.

\section{Failure cases}
We add a quantitative failure analysis on
ReasonVOS using Qwen2.5-VL-7B. Failures
($\mathcal{J}\&\mathcal{F}<0.5$; $149/458$ samples) are reported in the following table:

\begin{center}
\small
\begin{tabular}{lc}
\hline
First failing stage & \% of failures \\
\hline
Object extraction (LVLM) & 33.6 \\
Localization (adapter)   & 59.7 \\
Tracklet selection       & \phantom{0}6.7 \\
SAM2 propagation         & \phantom{0}0.0 \\
\hline
\vspace{-0.2cm}
\end{tabular}
\end{center}

Localization is the main bottleneck, accounting for $59.7\%$ of failures, followed by object extraction at $33.6\%$. Here, object extraction refers to whether the LVLM correctly identifies the target category implied by the query before localization begins. Localization failures occur when the correct category is identified but the adapted attention map does not sufficiently isolate the referred object. Tracklet selection contributes only $6.7\%$, indicating that once a reliable localization map is available, candidate ranking is generally robust. SAM2 is never the first failing stage, suggesting that propagation errors typically originate from an incorrect or ambiguous seed rather than from SAM2 itself.

\end{document}